# FSVL: Taxonomic Fast-Slow Vision-to-Language Collaborative Framework for Interpretable Confusable Fish Species Identification and Knowledge Generation in the Middle Yangtze River

Meng Liang[a,c†], Fucheng Wang[b†], Lang Zhao[b], Jun Chen[b], Ziying Li[b], Yuxiang Zhao[b], Haozhuang Chi[d], Wenkai Fang[e], Qiong Zhou[a], Junlin Yang[a], Jinming Wu[a*]

**Affiliations**

[a] *Key Laboratory of Freshwater Biodiversity Conservation, Ministry of Agriculture and Rural Affairs of China, Yangtze River Fisheries Research Institute, Chinese Academy of Fishery Sciences, Wuhan 430223, China*

[b] *College of Fisheries, Southwest University, Chongqing 400715, China*

[c] *College of Biosystems Engineering and Food Science, Zhejiang University, Hangzhou 310058, Zhejiang, China*

[d] *School of Mechanical and Aerospace Engineering, Nanyang Technological University, Singapore 639798, Singapore*

[e] *College of Computer Science and Technology, Zhejiang University, Hangzhou 310027, Zhejiang, China*

[†] Equal Contribution

[*]Corresponding Author：Jinming@yfi.ac.cn

**Abstract**

The Yangtze River Basin is a critical region for freshwater biodiversity conservation. The implementation of the Yangtze River Ten-Year Fishing Ban has increased the importance of effective aquatic biodiversity monitoring and species-level identification. However, fine-grained identification of closely related freshwater fish species remains challenging because of subtle morphological differences. Existing deep learning approaches often struggle to balance classification accuracy and inference efficiency, while large language models (LLMs) may generate unsupported information when producing specialized biological knowledge. To address these challenges, this study proposes FSVL, a fast-slow dual-pathway framework integrating visual recognition with LLM-based Retrieval-Augmented Generation (RAG). A lightweight ResNet50-based fast path performs preliminary classification across 51 fish species documented in Milddle Yangtze River Basin and routes predefined groups of easily confused species to a slow path for refined identification. Guided by taxonomic criteria, the slow path employs YOLOv8n to detect key morphological regions, including the head, dorsal fin, and tail, leveraging morphometric and outline characteristics to improve discrimination among closely related species.The framework further integrates a retrieval-enhanced backend based on Qwen3-14B, using soft token and Low-Rank Adaptation (LoRA) for parameter-efficient adaptation to domain-specific knowledge generation. Experimental results show that on the evaluated test set of 607 images covering Middle Yangtze River fish species, the standalone fast classifier achieved an accuracy of 99.67%, whereas the cascade successfully corrected the two errors, yielding an overall accuracy of 100.00%. These empirical findings indicate strong performance on the tested samples although further validation on independent specimens and external acquisition conditions remains necessary. In knowledge generation experiments involving 255 evaluation samples, scientific name accuracy increases from 91.37% to 100.00%, genus name accuracy improves from 90.20% to 98.82%, and ROUGE-L increases from 0.335 to 0.655. These results demonstrate the potential of FSVL to support fine-grained fish identification and domain-specific biological knowledge generation across the Yangtze River Basin, providing a technical approach for intelligent aquatic biodiversity monitoring and fisheries management.

## 1. Introduction

Freshwater fish biodiversity is an important indicator of aquatic ecosystem health and ecological integrity. As a critical freshwater ecosystem, the middle reaches of the Yangtze River host approximately 215 fish species, accounting for 60% of all freshwater fish species in the entire river system. Cyprinid-related groups dominate this region, and species differentiation relies heavily on subtle morphological traits, making visual identification extremely difficult (Shen et al., 2023). The implementation of the Yangtze River Ten-Year Fishing Ban has strengthened the need for reliable methods to monitor biodiversity recovery and support fisheries management. Accurate fish identification is essential for assessing changes in fish communities and enforcing conservation regulations. Non-destructive, vision-based monitoring provides a practical approach to identifying fish while minimizing disturbance to aquatic populations (Miller et al., 2025). However, distinguishing closely related species often requires specialized taxonomic expertise, which may not be readily available to frontline fishery administrators and conservation practitioners. Developing automated identification methods that combine accuracy, efficiency, and biological interpretability is therefore important for ecological monitoring in the middle reaches of the Yangtze River, aligning with broader methodological demands for scalable and transparent aquatic monitoring frameworks. Recent advances in artificial intelligence have highlighted the potential of automated species identification for large-scale biodiversity monitoring.

Despite advances in automated fish recognition, fine-grained identification of closely related species remains challenging (Wei et al., 2021; Veiga and Rodrigues, 2024). Many species exhibit subtle differences in head shape, mouth position, fin structure, scale patterns, and body markings. These characteristics can be difficult to distinguish under variations in posture, illumination, and image background (DeVries and Taylor, 2017). Conventional end-to-end classifiers primarily learn representations from overall image appearance, which may not adequately capture subtle local morphological differences(Zhang et al., 2026). While attention mechanisms such as CBAM can adaptively highlight salient features (Woo et al., 2018), they do not explicitly encode taxonomic criteria, limiting their biological interpretability. Increasing model complexity can improve discrimination but may also increase computational costs (Canziani et al., 2017). Moreover, many visual classification systems do not explicitly incorporate taxonomic criteria, limiting the biological interpretability of their predictions. A further challenge is the generation of reliable biological knowledge from recognition results. LLMs can produce domain-specific descriptions, but their outputs may contain unsupported morphological or ecological information. Retrieval-Augmented Generation (RAG) can ground responses in external knowledge, although retrieval alone does not guarantee consistency between generated content and retrieved evidence (Lewis et al., 2021). These limitations motivate the development of an integrated framework for efficient recognition, taxonomically informed discrimination, and reliable knowledge generation (Veiga and Rodrigues, 2024; Mohammadisabet et al., 2025).

Traditional fish identification relies on expert examination of morphological characteristics and comparison with established taxonomic criteria. Although scientifically grounded, this approach requires specialized expertise and can be time-consuming for large-scale monitoring. Machine learning and convolutional neural networks, largely driven by large-scale datasets like ImageNet have improved the efficiency of image-based fish classification, but distinguishing closely related species remains difficult when diagnostic differences are concentrated in small anatomical regions. Object detection and local feature extraction can focus on informative structures, while combining local and global features can provide complementary morphological representations (Woo et al., 2018). Nevertheless, visual classification alone does not necessarily provide taxonomic explanations or structured biological knowledge. LLMs and RAG offer opportunities to connect visual recognition with specialized knowledge bases (Janati et al., 2026), but factual consistency remains a concern. Parameter-efficient adaptation methods, including soft token and LoRA, provide potential approaches for adapting language models to specialized generation tasks without updating all model parameters (Han et al., 2024). An effective framework should therefore coordinate visual recognition and knowledge generation, allocate additional computation to difficult cases, and generate biological information grounded in retrieved evidence.

To address these challenges, this study proposes FSVL, an expert-guided fast-slow collaborative framework for fine-grained fish identification and knowledge generation (Fig. 1). First, a ResNet50-based fast path performs preliminary classification across 51 fish species, while a computationally intensive slow path is activated for pre-defined groups of easily confused species. Second, a YOLOv8n-based detector locates diagnostic morphological regions, including the head, dorsal fin, and tail. Local features are combined with global image representations to improve discrimination among closely related species. Third, the backend integrates a fish knowledge base with the Qwen3-14B through RAG (Carion et al., 2025). Soft token and LoRA are used for parameter-efficient model adaptation, while a response-only loss strategy focuses training on generated answers rather than input prompts. The main contributions are threefold. First, a fast-slow collaborative architecture allocates additional computation to predefined groups of confusing species. Second, a taxonomically guided global-local feature fusion strategy combines diagnostic regions with whole-image information for fine-grained recognition. Third, a knowledge-generation module integrates an external fish knowledge base with parameter-efficient LLM adaptation. The framework is evaluated on a dataset of 51 fish species from the middle reaches of the Yangtze River, with experiments assessing both recognition performance and the quality of generated biological knowledge.

## 2. Materials and Methods

The overall architecture of the proposed FSVL framework is illustrated in Fig. 1. First, input specimen images are standardized via SAM3-based background removal, after which a shared ResNet50 encoder performs rapid preliminary classification across all 51 fish species. If the prediction falls into predefined confusable taxa ( $y_{\text{fast}} \in \mathcal{C}_{\text{slow}}$ ), the slow pathway is triggered to fuse global and local anatomical features (head, dorsal fin, and tail) for fine-grained discrimination. Finally, the confirmed species label guides a RAG-enhanced Qwen3-14B model (adapted with soft token and LoRA) to generate structured biological profiles.

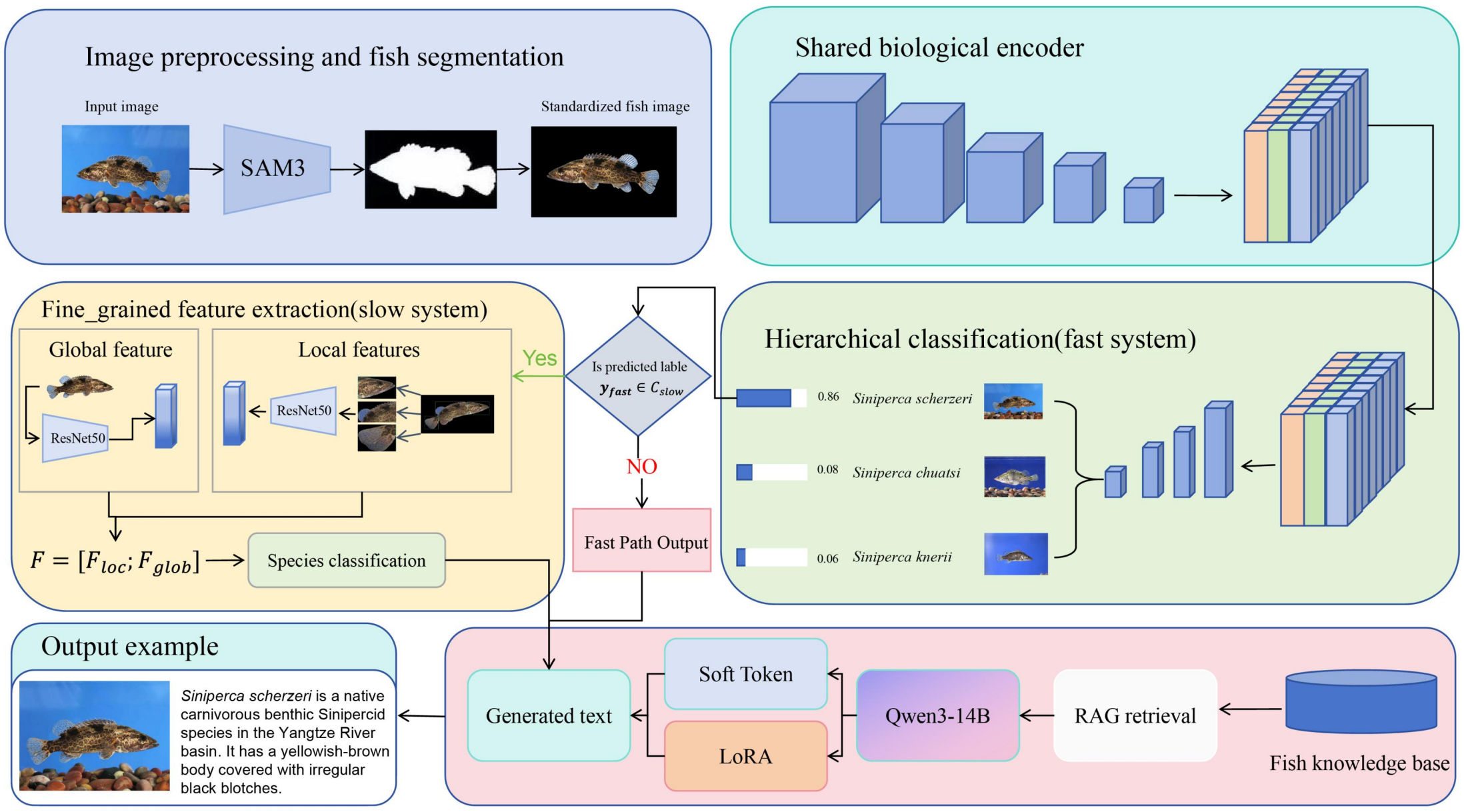

**Fig. 1.** Overall architecture of the fast to slow dual path fish recognition system with RAG knowledge enhancement.

## *2.1 Dataset Construction*

This study constructed an image dataset consisting of 51 typical freshwater fish species from Middle Yangtze River, covering 6 orders, 13 families, and 30 genera of ray finned fish. It encompasses the main economic fish species and rare protected species in this region. The data collection locations included Yichang, Xiaoting, Jingzhou, Shishou, and Wuhan sections of Middle Yangtze River (Fig. 2). Field sampling involved close collaboration with local fishermen and recreational anglers along the riverbank. To ensure high fidelity and standardized specimen documentation, live or freshly sampled fish specimens were photographed under controlled indoor and outdoor aquarium conditions using professional DSLR cameras (Canon EOS 5D Mark II and Canon EOS 5D Mark IV) equipped with a macro lens (Tamron SPA AF 90mm f/2.8 Di Macro) and external diffused flash units (Canon Speedlite 430EX III RT) to eliminate specular reflections and maintain consistent illumination across specimens. As illustrated in Fig. 2(c) and (d), a specially designed modular photographic setup comprising front, middle, and rear optical shielding curtains, a supporting table, a glass tank with a partition, and calibrated background control, was employed to isolate the fish specimens from environmental noise and standardize visual inputs.

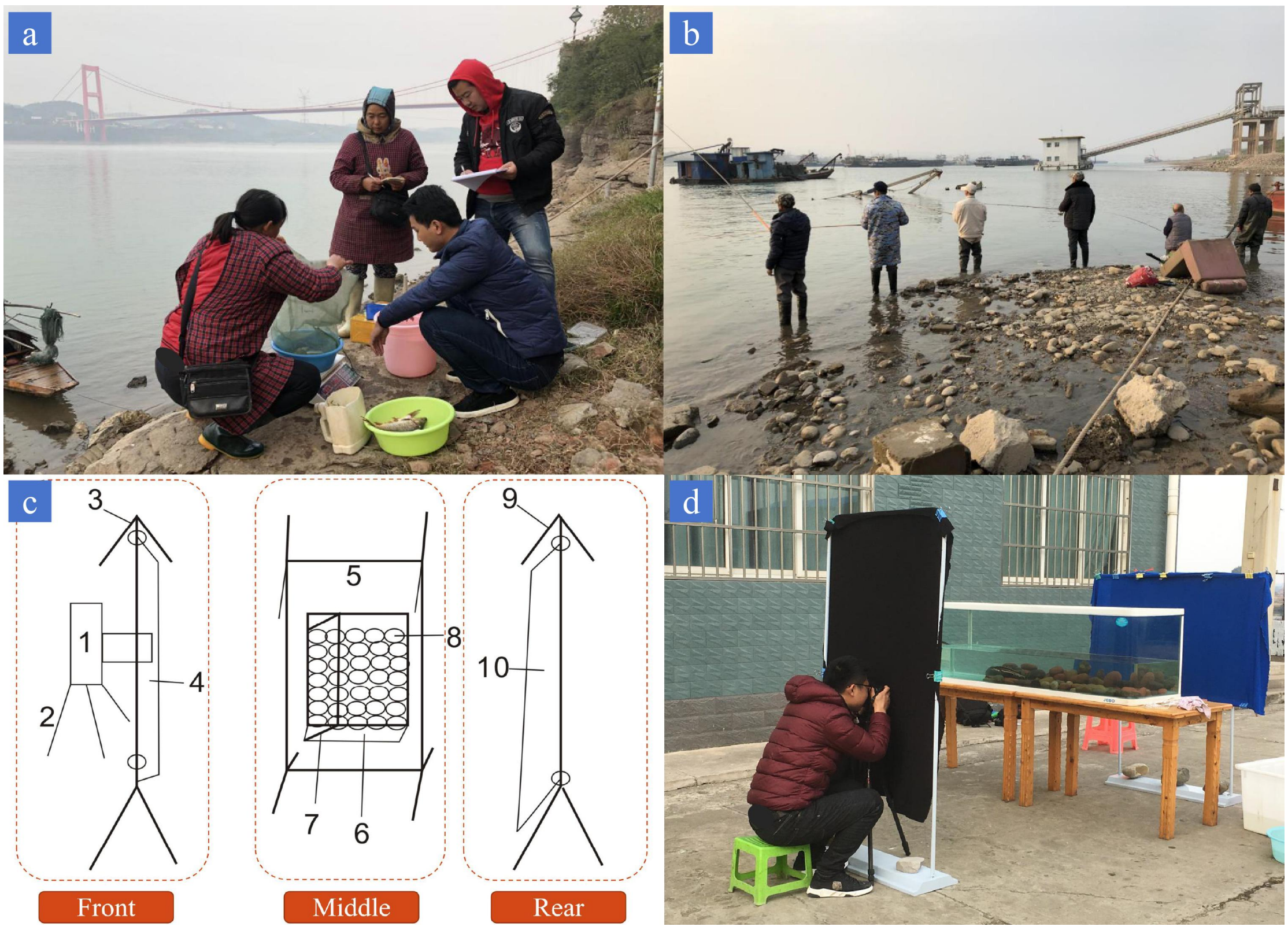

**Fig. 2.** Field sample collection methods and live fish photographic setups from the middle reaches of the Yangtze River. (a) Acquisition of freshly caught fish samples from local fishermen after landing; (b) Procurement of fish specimens from recreational anglers along the riverbank; (c) Schematic diagram of the live fish photographic setup (Front, Middle, and Rear views): 1, camera; 2, 3, and 9, stands; 4 and 10, curtains; 5, supporting table; 6, glass tank; 7, glass partition; 8, pebbles; (d) On site photograph showing the live fish photography process in a controlled environment.

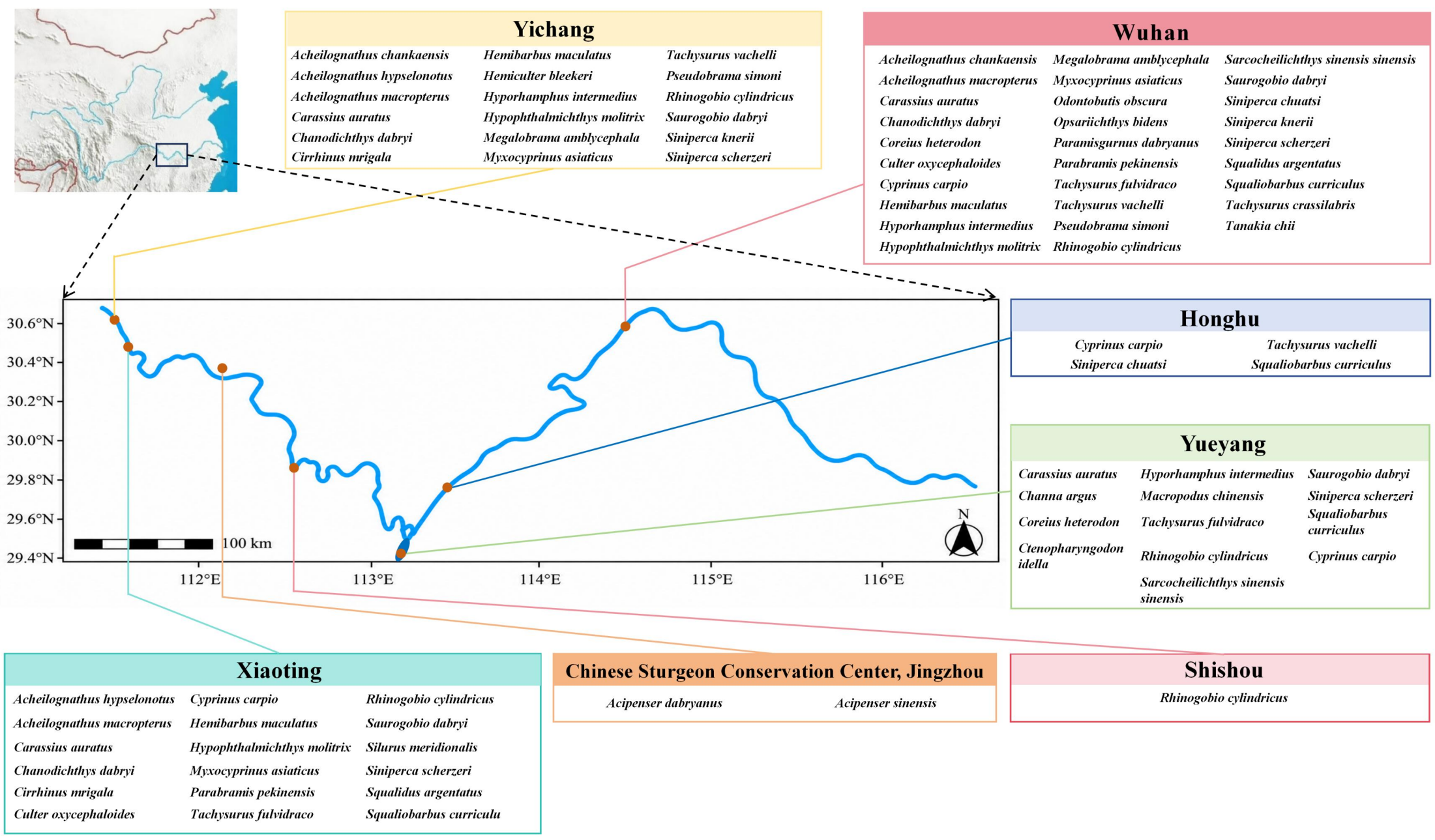

**Fig. 3.** Distribution of fish sampling sites along the middle Yangtze River basin.

By collecting and organizing the original images, high definition images of each type of fish were captured. The folders were organized according to the naming conventions of the family, genus and species, and an initial dataset consisting of 51 categories and approximately 6,900 original images was constructed. The majority of fish had between 50 and 150 original images, while a few species had about 10 images, and some species had around 250 images (Fig. 4).

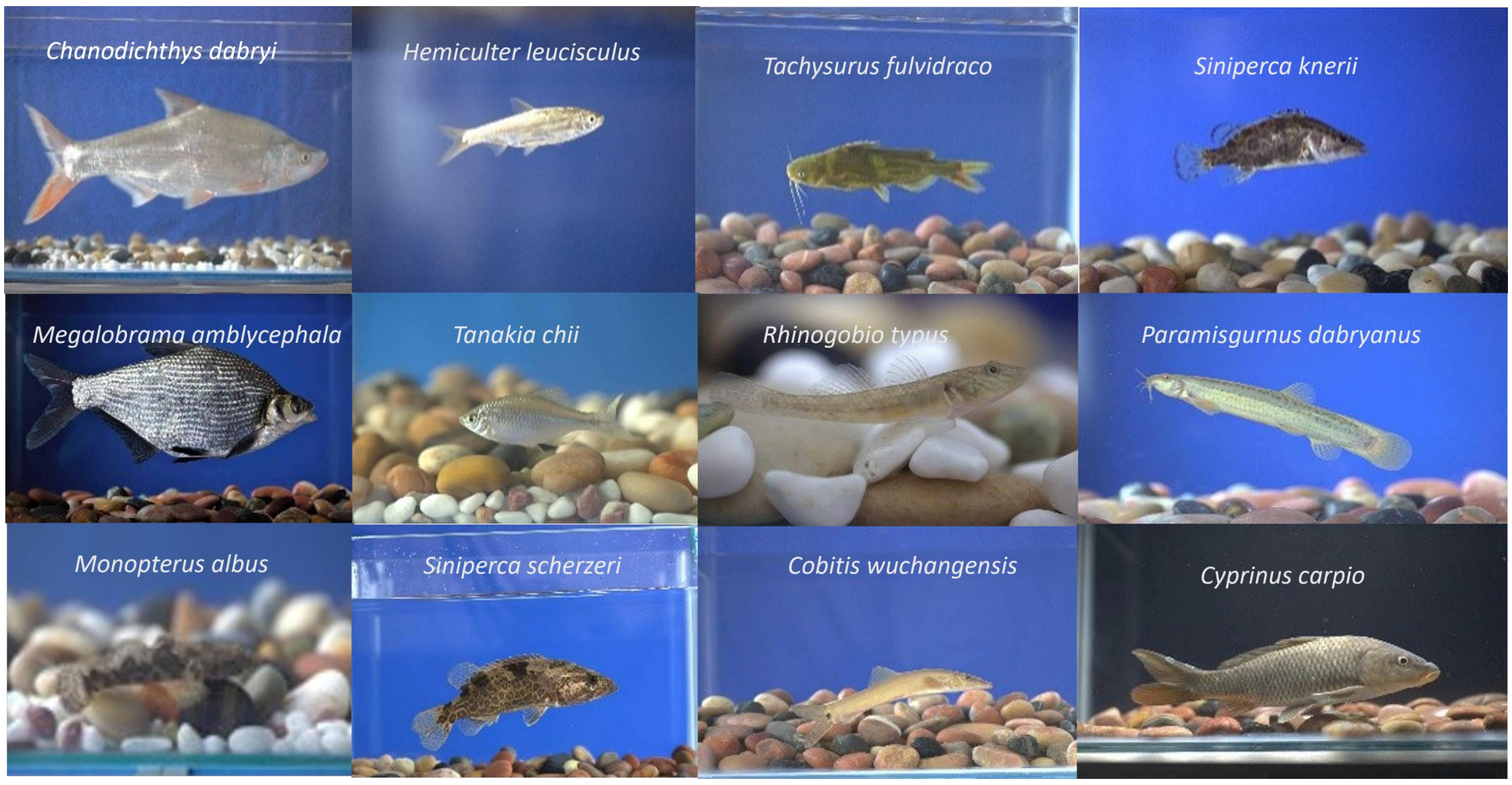

**Fig. 4.** Partial dataset and sample images.

Then, data augmentation and division were carried out. To ensure the statistical reliability of the evaluation results for the test set, data augmentation operations were conducted for the categories with fewer than 80 original images. The operations included random rotation, horizontal flipping, brightness and contrast adjustment, etc., so that the number of images in each category reached more than 80. Subsequently, the dataset was divided into training set, validation set, and test set in a ratio of 8:1:1. To ensure rigorous evaluation and prevent data leakage, individual-level data partitioning was enforced, ensuring that all images originating from the exact same individual fish specimen were strictly assigned to the same subset. Data augmentation was applied exclusively to the training set to balance minority classes, while independent test images remained completely unaugmented. After background removal was performed, and all images were automatically segmented using the SAM3 model to remove background interference and output transparent background images of fish bodies, which were used as input data for model training (Carion et al., 2025). Specifically, the resulting dataset comprises 4,818 training images for model optimization and an independent test set of 607 images for the joint evaluation of the fast and slow pathways. In the large model experiment, the division was carried out according to the fish species records to ensure that the test fish species did not appear in the training set, in order to evaluate the generative generalization ability of the model for unseen fish species. The data division ratios and specific quantities of each part of the experiment vary, and detailed information can be found in Section 2.6 of the experimental settings.

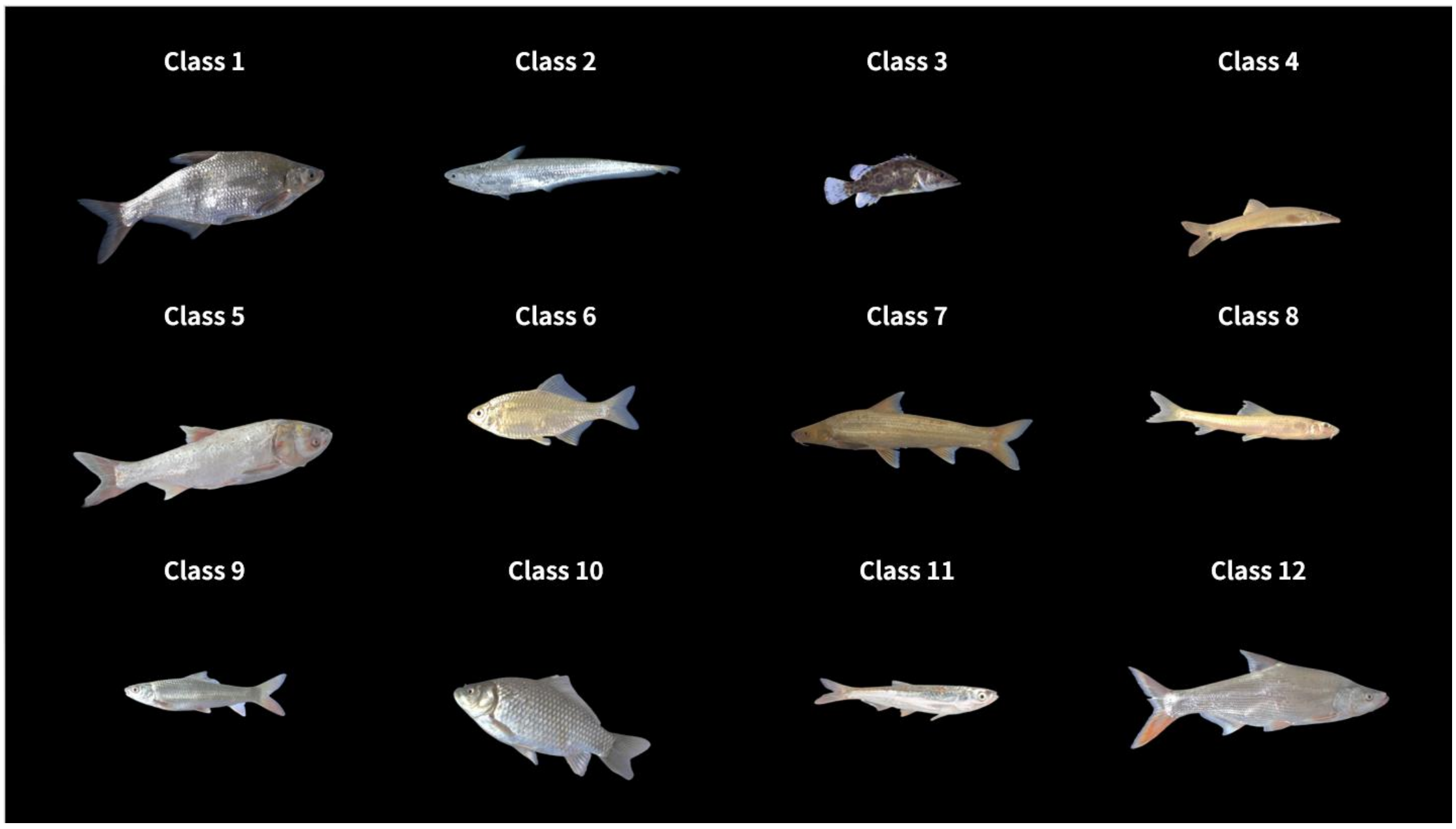

**Fig. 5.** Examples of fish specimen images after background removal by SAM3.

To ensure comprehensive ecological representation, the constructed dataset encompasses not only dominant economic species but also representative threatened, endangered, and nationally protected fish species native to the middle reaches of the Yangtze River, as detailed in Table 1.

**Table 1**

Representative threatened, endangered, and nationally protected fish species inhabiting Middle Yangtze River. *Acipenser schrenckii* specimens were photographed from Chinese Sturgeon Conservation Center, Jingzhou.

| Scientific Name | IUCN Red List Status | China National Protection Level | CITES Appendix |
|---|---|---|---|
| *Acipenser sinensis* | Critically Endangered (CR) | Class I | Appendix II |
| *Acipenser dabryanus* | Extinct in the Wild (EW) | Class I | Appendix II |
| *Myxocyprinus asiaticus* | Vulnerable (VU) | Class II | Not Listed |
| *Acipenser schrenckii* | Critically Endangered (CR) | Class II (Wild populations only) | Appendix II |

Note: IUCN status abbreviations: EW, Extinct in the Wild; CR, Critically Endangered; VU, Vulnerable.; DD, Data Deficient. Evaluations follow the International Union for Conservation of Nature Red List criteria.

### *2.2 Key site definition guided by expert taxonomy*

The traditional morphological classification of fish mainly relies on countable morphological features and measurable morphological features. The standardized visual identification process by field experts usually follows four steps: first, narrowing the range by observing body color and patterns, then, determining the family by inspecting the head and mouth position, subsequently, establishing the genus level classification based on fins and pelvic spines, and finally, completing the species identification through precise counting of the lateral line and scales. Representative examples of expert diagnostic criteria for typical confusable species pairs are illustrated in Table 2. Inspired by this taxonomic paradigm, this study selects three key anatomical regions, the head, dorsal fin, and tail, as the local feature extraction areas for the slow path, explicitly encoding expert taxonomic methodology into the model architecture.

**Table 2**

Key morphological diagnostic characteristics of typical closely related fish species in the middle reaches of the Yangtze River.

| Fish species are paired | Observation dimensions | Fish species A feature description | Fish species B feature description | Experts quickly identify the criteria |
|---|---|---|---|---|
| *Coilia nasus* vs. *Coilia brachygnathus* | Upper jaw length, number of lateral line scales, number of anal fin rays | *Coilia nasus*: The maxilla is very long, extending well beyond the gill opening to the base of the pectoral fin or beyond. Lateral line scales: 74–84; anal fin rays: 74–89. | *Coilia brachygnathus*: The maxilla is relatively short, extending only to or slightly beyond the gill opening and not reaching the base of the pectoral fin. Lateral line scales: 68–78; anal fin rays: 72–82. | Check whether the posterior end of the maxilla reaches or extends beyond the pectoral fin base. A longer maxilla supports *Coilia nasus*, whereas a shorter maxilla supports *Coilia brachygnathus*. |
| *Megalobrama amblycephala* vs. *Megalobrama terminalis* | Mouth cleft shape, standard length-to-body depth ratio, branched anal fin ray count | *Megalobrama amblycephala*: Mouth terminal, with the mouth cleft nearly horizontal; body deep and broadly oval (standard length 2.0 – 2.3 times body depth); branched anal fin rays: 27–32. | *Megalobrama terminalis*: Mouth terminal, with the mouth cleft distinctly oblique; body relatively less deep and diamond shaped (standard length 2.2 – 2.7 times body depth); branched anal fin rays: 24–28. | Compare mouth cleft orientation and body depth. *Megalobrama amblycephala* has a more nearly horizontal mouth cleft and a deeper body, whereas *Megalobrama terminalis* has an oblique mouth cleft and a relatively less deep body. Branched anal fin ray counts provide supplementary evidence. |
| *Culter alburnus* vs. *Culter mongolicus* | Mouth cleft orientation, lateral line scale count, and coloration of the lower lobe of the caudal fin | *Culter alburnus*: Mouth superior, with the mouth opening nearly vertical relative to the body's longitudinal axis and the lower jaw strongly upturned. Lateral line scales: 80 to 94; caudal fin grey. | *Culter mongolicus*: Mouth superior, but the mouth cleft is oblique rather than vertical; the lower jaw projects forward but is not vertically upturned. Lateral line scales: 73–77; the lower lobe of the caudal fin is bright red. | Compare mouth cleft orientation and the coloration of the lower caudal fin lobe. A nearly vertical mouth opening supports *Culter alburnus*, whereas an oblique mouth cleft and a bright red lower caudal fin lobe support *Culter mongolicus*. Scale counts provide supplementary evidence. |
| *Siniperca chuatsi* vs. *Siniperca knerii* | Marking pattern, dorsal curvature (body depth), pyloric caeca count | *Siniperca chuatsi*: Body with large, irregular, cloud like dark brown patches; dorsal profile strongly arched, forming a pronounced hump. Pyloric caeca: approximately 100–200. | *Siniperca knerii*: Body, head, and fins densely covered with small, rounded or oval black spots; body relatively slender, with a gently arched dorsal profile. Pyloric caeca: reported as approximately 300–400. | Distinguish the large, irregular dark patches of *Siniperca chuatsi* from the numerous small, rounded or oval spots of *Siniperca knerii*. Body profile provides additional evidence. Pyloric caeca counts require anatomical examination. |
| *Coreius heterodon* vs. *Coreius guichenoti* | Snout shape, relative eye diameter, maxillary barbel length, and lateral line scale count | *Coreius heterodon*: Snout relatively pointed, with comparatively larger eyes. The pair of maxillary barbels is relatively short and does not reach the pectoral fin base. Lateral line scales: 54–56. | *Coreius guichenoti*: Snout broad, rounded, and blunt, with relatively small eyes. The maxillary barbels are long, extending beyond the pectoral fin base. Lateral line scales: 55–58. | Compare snout shape, relative eye size, and maxillary barbel length. A pointed snout and shorter barbels support *Coreius heterodon*; a blunt, rounded snout, smaller eyes, and longer barbels support *Coreius guichenoti*. Lateral line scale counts overlap and should not be used alone. |

*2.3 Dual path Fast Slow Collaboration*

Architecture Based on the above expert classification knowledge, this study has constructed a dual path fast slow collaborative recognition architecture, aiming to balance the efficiency and accuracy of recognition. The system adopts a pipeline design of shared pre-processing with dynamic routing, mainly consisting of four core modules. Firstly, the shared pre-processing layer processes the input image using the SAM3 to complete automated image segmentation (Carion et al., 2025), removing background interference and outputting a standardized fish image with a size of 224×224 pixels while accounting for optical degradations in complex aquaculture environments (Yang et al., 2026) Then it enters the fast path recognition module, using ResNet50 as the main backbone network (He et al., 2016), training on a 51 class fish dataset to achieve rapid preliminary classification. The fast path outputs the predicted category, which serves as the basis for subsequent routing decisions. Finally, it enters the slow path fine recognition module. When the predicted category by the fast path belongs to the pre-defined 6 types of easily confused closely related species, the system activates the slow path. This module consists of a YOLOv8n part detector, a global local feature fusion module, and a fusion classifier. The YOLOv8n part detector mainly detects and boxes the three key parts of the fish body: the head, dorsal fin, and tail, incorporating attention-enhanced feature representations tailored for underwater vision (Hamzaoui et al., 2026).The global local feature fusion module extracts 2048 dimensional global features and 2048 dimensional local features for each part, concatenating them into an 8192 dimensional fused vector, and the fusion classifier classifies the fused vector to output the species label after fine recognition. The dynamic routing mechanism is formulated as Eq.(1) to Eq.(3).

$$\text{Route}(\mathbf{x}) = \begin{cases} \text{Slow Path}, & \text{if } \arg\max(f_{\text{fast}}(\mathbf{x})) \in C_{\text{slow}} \\ \text{Fast Path}, & \text{otherwise} \end{cases} \tag{1}$$

$$\mathcal{L}_{\text{fast}} = -\frac{1}{N}\sum_{i=1}^{N}\sum_{c=1}^{C} y_{i,c} \log\left(\frac{\exp(z_{i,c})}{\sum_{j=1}^{C}\exp(z_{i,j})}\right) \tag{2}$$

$$\hat{y}_{\text{fast}} = \arg\max_{c\in\{1,\ldots,51\}} P(y=c \mid x) \tag{3}$$

Among them, $C_{slow}$ represents a predefined set of 6 groups of easily confused closely related species. This set was determined based on the confusion matrix analysis of the fast path on the validation setthe categories with the highest error rates identified by the fast path were selected as the coverage range of the slow path. These confusable species include *Hemiculter leucisculus, Pseudobrama simoni, Squalidus* *argentatus, Acheilognathus hypselonotus, Tachysurus crassilabris,* and *Tachysurus vachellii.*

**Table 3**

Top confusable fish species pairs and their classification error statistics evaluated across the 51 species dataset.

| True Species (Slow Path Classes) | Misclassified As (predicted Species) | Morphological Similarity & Confusion Reason |
|---|---|---|
| *Hemiculter leucisculus* | *Squalidus argentatus / Pseudobrama simoni* | Con-generic or closely related cyprinids sharing extremely similar body elongation, silver-white flank coloration, and overlapping lateral line scale counts. |
| *Acheilognathus hypselonotus* | *Hemiculter leucisculus* | High-bodied lateral profiles and similar vertical banding patterns during juvenile or compressed developmental stages leading to fast-path confusion. |
| *Tachysurus crassilabris / Leiocassis crassilabris* | *Pelteobagrus vachellii* | Benthic bagrid catfishes sharing comparable body shapes, dark brownish pigmentation, and tactile whisker configurations. |

**Algorithm 1: Fast Slow Dual Pathway Inference for Fish Identification**

**Input**:
Raw image $I \in R^{H\times W\times 3}$
Predefined confusable species set C_slow Fine-Grained
Fast path classifier M_fast, YOLOv8n detector D_YOLO
Global and local feature extractors Φ_g, Φ_l
Fusion classifier F_fusion

**Output**:
Predicted species label y_hat
Inference pathway indicator s ∈ {"fast", "slow"}

```
x ← x ← (Resize(I) − μ) / σ                      // Resize to 224×224 and normalize
p_fast ← Softmax(M_fast(x))                      // Output probability distribution
ŷ_fast ← arg max_{c∈C} p_fast(c)_{c ∈ C} p_fast(c)   // Initial classification
if y_hat_fast ∈ C_slow then
    {I_head, I_dorsal, I_tail} ← D_YOLO(x)      // Detect and crop key parts
    if |Parts| == 3 then
        F_g ← Φ_g(x)                             // Global feature, F_g ∈ R^2048
        F_head ← Φ_l(I_head)
        F_dorsal ← Φ_l(I_dorsal)
        F_tail ← Φ_l(I_tail)                     // Local features, F_{·} ∈ R^2048
        F_fusion ← [F_g; F_head; F_dorsal; F_tail]   // F_fusion ∈ R^8192
        F_norm ← (F_fusion − μ) / sqrt(σ^2 + ϵ)
        ŷ_slow ← arg max_{k∈C_slow} Softmax(W_3 · ReLU(W_2 · ReLU(W_1 F_norm + b_1) + b_2) + b_3)_k
```

```
            return y_hat_slow, "slow"
        end if
end if
return y_hat_fast, "fast"
end
```

Furthermore, to quantitatively evaluate the system-level computational efficiency under cascaded execution, let $p$ denote the empirical proportion of samples triggering the slow path during field operation; the system-level average inference latency follows a cascaded pipeline formulation in Eq. (4). It should be noted that Table 8 outlines the isolated computational cost of the visual classification backbones and does not yet incorporate the cumulative overheads of SAM3 image segmentation, YOLOv8n local region detection, external knowledge retrieval, and Qwen3-14B text generation. Consequently, future deployments must rigorously differentiate between 'image-to-taxon recognition latency' and 'image-to-knowledge report generation latency' across diverse hardware configurations:

$$T_{avg} = T_{preprocess} + T_{fast} + p \cdot T_{slow,additional} \tag{4}$$

Here, $T_{\mathrm{fast}}$ and $T_{slow,additional}$ represent the single inference time for the fast path and the slow path respectively. It should be noted that this formulation outlines the isolated computational cost of the visual classification backbones and does not yet incorporate the cumulative overheads of SAM3 image segmentation, YOLOv8n local region detection, external knowledge retrieval, and Qwen3-14B text generation. Consequently, future deployments must rigorously differentiate between 'image-to-taxon recognition latency' and 'image-to-knowledge report generation latency' across diverse hardware configurations.

### *2.4 Global Local Feature Fusion*

The core of the slow path lies in integrating global and local features (Prasetyo et al., 2022; You et al., 2026; Zhang et al., 2026) to capture the subtle morphological differences among closely related species. The feature fusion process is as follows: Firstly, global feature extraction is conducted. A complete fish body image of size 224 × 224 is input into the global encoder (ResNet50), and after removing the classification head, a 2048 dimensional global feature vector is extracted, which is used to represent information such as the fish's body shape, color, and overall outline. Then, local feature extraction is carried out. Based on the core identification criteria of fish taxonomy, the head, dorsal fin, and tail are located using the YOLOv8n detector and local images are cropped, reflecting part-based anatomical segmentation and feature extraction principles (Feng et al., 2023).The three cropped local regions are processed by a shared local ResNet50 encoder (initialized with ImageNet pre-trained weights, distinct from the global encoder), and three 2048 dimensional local feature vectors are extracted. As defined in Eq. (5) to Eq. (8), feature fusion combines the global features with the three local features to form an 8192 dimensional fused vector.

$$F_{\text{concat}} = [F_{\text{global}} \oplus F_{\text{head}} \oplus F_{\text{dorsal}} \oplus F_{\text{tail}}] \in \mathbb{R}^{8192} \tag{5}$$

$$h_2 = \text{ReLU}(W_2 h_1 + b_2) \tag{6}$$

$$\hat{y}_{\text{slow}} = \text{Softmax}(W_3 h_2 + b_3) align * \tag{7}$$

$$\mathbf{F}_{\text{fusion}} = [\mathbf{F}_g ; \mathbf{F}_h ; \mathbf{F}_d ; \mathbf{F}_t] \in \mathbb{R}^{8192} \tag{8}$$

The fused vectors are then input into the fusion classifier for the final judgment. This design enables the model to simultaneously focus on the overall outline of the fish body and the key local details, effectively capturing the extremely subtle morphological differences among closely related species.

### *2.5 RAG Knowledge Enhancement Module*

The system backend integrates a knowledge enhancement module based on RAG, aligning with recent advances in retrieval-augmented multimodal frameworks for species identification (Asai et al., 2024; Wang, 2026). Upon receiving species classification labels, the system queries a prebuilt knowledge base containing structured data on morphological features, ecological habits, and distribution information across the 51 fish species. Retrieved evidence is subsequently provided as contextual input to the large language model to generate structured biological descriptions.

To suppress hallucination risks and ensure domain adaptability, the framework freezes the core parameters of Qwen3-14B while optimizing lightweight adaptation components, specifically combining soft token and LoRA (Yang et al., 2025). Soft Token introduces learnable continuous prompt vectors concatenated at the input sequence to guide knowledge organization, whereas LoRA injects trainable rank decomposition matrices into the attention weights. Furthermore, training employs a masked response-only loss strategy, restricting loss computation exclusively to generated answers while masking input prompts, thereby enabling efficient parameter adaptation without altering foundational weights. The input sequence adaptation with soft token, the internal weight adaptation via LoRA, and the joint parameter optimization objective under the masked response-only loss are formulated in Eq. (9), Eq. (10), and Eq. (11), respectively.

$$H_0' = [S; \text{Embedding}(X)] \tag{9}$$

$$W' = W + BA \tag{10}$$

$$\min_{S,A,B} \mathcal{L}_{\text{gen}} = -\sum_{t=1}^{T} \log P_{\Theta,\Phi,S}(y_t \mid R, X, y_{<t}) \tag{11}$$

Among them, $\mathrm{x} = [\mathrm{RAG}, \mathrm{Instruction}], \mathrm{y} = \mathrm{Answer}$. This design enables the model to focus on learning how to generate answers based on retrieved knowledge, rather than relearning the input prompt itself. Therefore, soft token and LoRA jointly achieve efficient and effective adaptation of the joint parameters in the input representation space and the model parameter space. RAG is responsible for providing factual basis, soft token is responsible for learning the continuous input prompts related to the task, and LoRA is responsible for lightweight adaptation of the attention weights within the model.

### *2.6 Experimental Setup*

**Table 4**

Experimental parameter settings for each module.

| Module | Backbone / Model | Epochs | Batch size | Learning rate | Optimizer | Other settings |
|---|---|---|---|---|---|---|
| Fast path (Stage 1) | ResNet50 | 15 | 32 | 0.001 | Adam | Flip, rotation (±10° ) |
| slow path (Stage 2) | ResNet50 | 5 | 32 | 1e-5 | Adam | Color Jitter (0.2), flip, rotation |
| YOLO detector | YOLOv8n | 100 | 8 | Default | Default | Input 640 × 640, mAP50 = 0.98 |
| Slow path fusion classifier | 3layer FC | 100 | 64 | 0.001 | Adam | Dropout = 0.3, input 8192 D |
| Backend LLM | Qwen3-14B | See Supplementary Material | — | — | — | Soft token = 8, LoRA rank = 8 |

The fast path utilizes ResNet50 as the backbone network, initialized with ImageNet pre-trained weights. Model optimization is conducted across two distinct stages. The first stage trains for 15 epochs on the 51-class training set with an initial learning rate of 0.001, batch size of 32, Adam optimizer, cross-entropy loss, and a gradient clipping threshold of 1.0. The second stage loads weights from the first stage and continues training for 5 epochs with a reduced learning rate of 1e-5 alongside Color Jitter data augmentation to enhance feature discrimination.

The YOLO detector is trained on annotated data covering the six confusable congeneric species, utilizing an input resolution of 640×640 pixels, batch size of 8, and 100 training epochs on an NVIDIA RTX 4080 Super GPU. The slow path fusion classifier is implemented as a three-layer fully connected network processing 8192-dimensional concatenated features, trained for 100 epochs with the Adam optimizer at a learning rate of 0.001. Finally, backend large language model adaptations are carried out using Qwen3-14B with 8 learnable soft prompt vectors and LoRA rank of 8, completing an integrated pipeline from visual classification to knowledge generation.

## 3. Results

### *3.1 Fast Path Results*

The training convergence and recognition performance of the fast path evaluated on the 51 species training set 4,818 images are shown in Table 5. After the first stage of training, the training evaluation accuracy of the fast path reached 96.72%. After the second stage of fine-tuning learning rate $1\times10^{-5}$, 5 epochs with Color Jitter data augmentation, the accuracy on the 4,818 training samples increased to 99.29%, representing an improvement of 2.57 percentage points furthermore, when evaluated on the independent 607 image test set, the two-stage fast path achieved a standalone accuracy of 99.67%, as detailed in Section 3.2.2. This indicates that on a high-quality dataset with background segmentation, reducing the learning rate and introducing Color Jitter augmentation enable the model to further converge in the later stage, significantly improving its ability to distinguish closely related fish species.

**Table 5**

Recognition accuracy of the fast path across the two training stages evaluated on the 4,818 image training set.

| Training stage | Learning rate | Epochs | Accuracy |
|---|---|---|---|
| Stage 1 | 0.001 | 15 | 96.72% |
| Stage 2 | 1×10-5 | 5 | 99.29% |

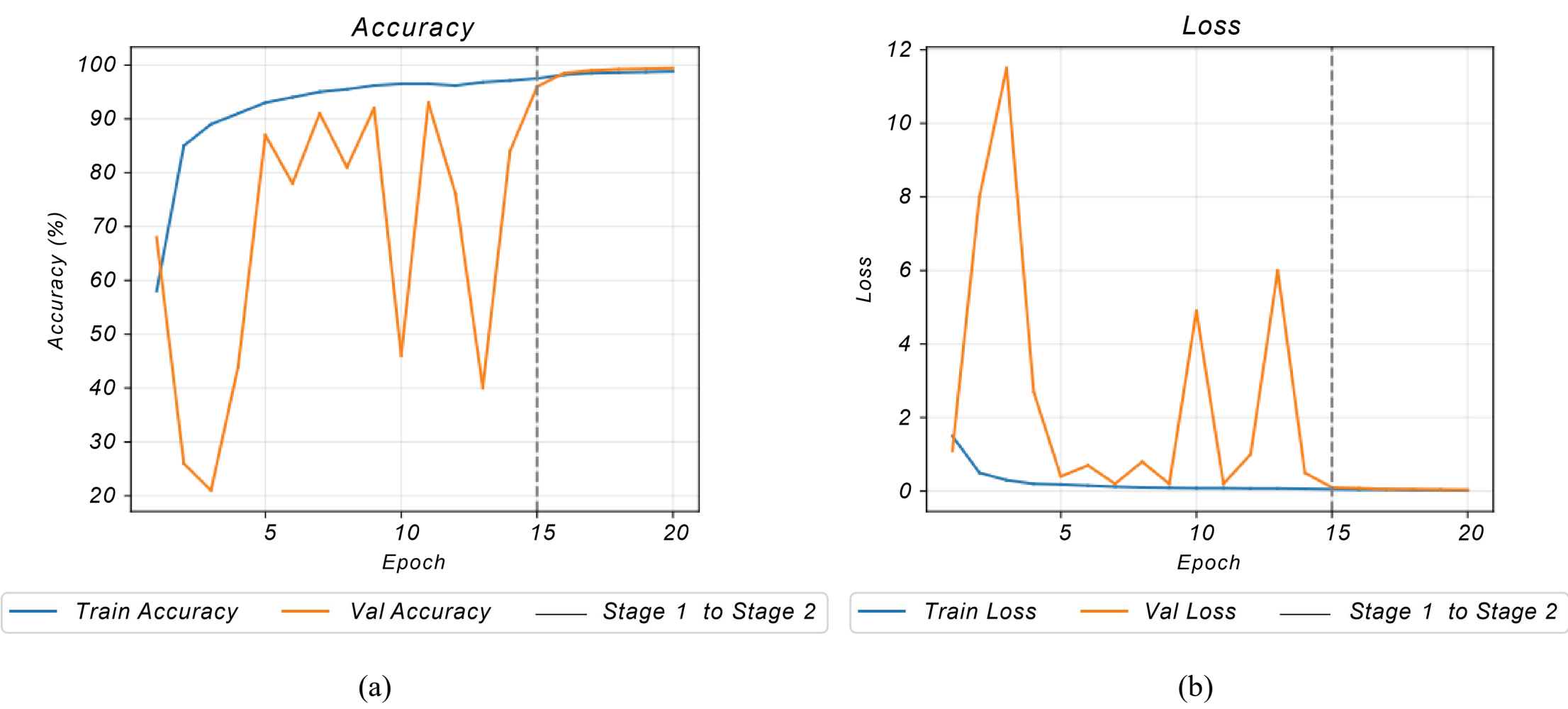

**Fig. 6.** Fast path training curve. (a) Training and validation accuracy varies with epoch, (b) Training and validation loss varies with epoch. The dashed line represents the boundary between the first and second stages.

### *3.2 Slow Path Results*

#### *3.2.1 Key Part Localization Results of the YOLO Detector*

The slow path consists of the YOLO key part detector and the global to local feature fusion classifier. It is activated only when the fast path prediction results fall into the 6 types of easily confused close species. For the 6 types of easily confused close fish species, we labelled the three key parts (head, dorsal fin, and tail) for each image and trained the YOLO detector for automatic localization. The labelled data was divided into training set, validation set, and test set in a ratio of 8:1:1. The YOLO detector achieved excellent detection performance on the validation set 135 images, 405 instances. The overall mAP50 reached 0.98, and mAP50 95 reached 0.804. From each category, the precision of the head was 0.973, the recall was 0.985, the mAP50 was 0.979, and mAP50 95 was 0.774. The precision of the dorsal fin was 0.977, the recall was 0.958, the mAP50 was 0.967, and mAP50 95 was 0.792. The precision of the tail was 0.985, the recall was 0.996, the mAP50 was as high as 0.995, and mAP50 95 was 0.847. The tail detection accuracy was the highest because the shape features of the tail fin are obvious and the boundaries are clear, making it easy for the model to accurately capture. The mAP50 95 of the head and dorsal fin were slightly lower than that of the tail, but the mAP50 was above 0.96 for both, indicating that the bounding box localization was basically accurate. In terms of inference speed, the average inference time for a single image was only 0.8 ms. Including preprocessing and postprocessing, the total time was approximately 2.0 ms. The model size was only 6.2 MB, making it easy to deploy on edge devices. The detection results are shown in Table 6.

**Table 6**

Detection performance of the YOLO detector for each key part.

| Class | Precision | Recall | mAP50 | mAP50 95 |
|---|---|---|---|---|
| Head | 0.973 | 0.985 | 0.979 | 0.774 |
| Dorsal fin | 0.977 | 0.958 | 0.967 | 0.792 |
| Tail | 0.985 | 0.996 | 0.995 | 0.847 |
| Overall | 0.979 | 0.98 | 0.98 | 0.804 |

Note: The validation set contains 135 images with 405 instances. The inference speed is 0.8 ms per image for detection, and approximately 2.0 ms including pre- and post-processing. The model size is 6.2 MB.

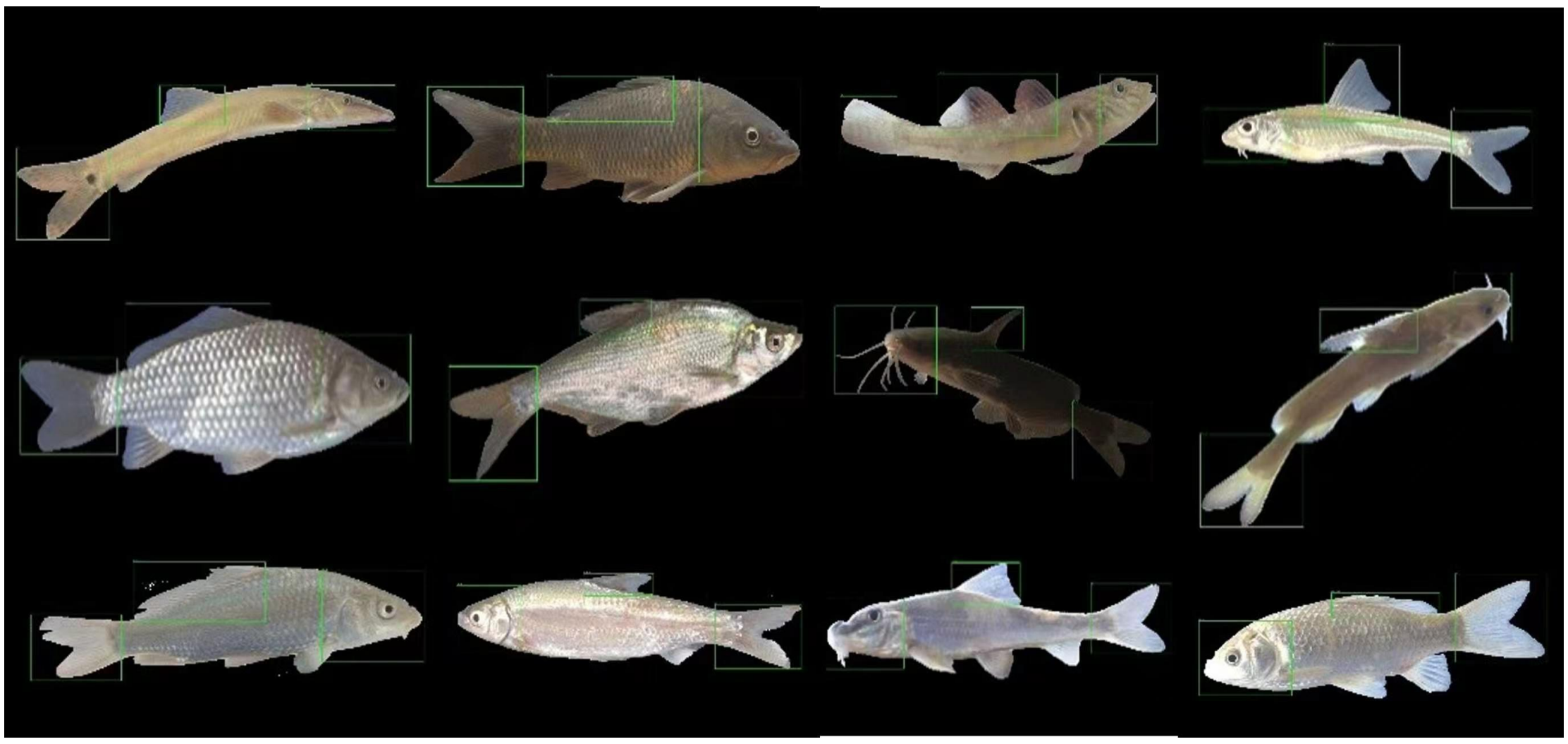

**Fig. 7.** Example output from YOLOv8n detector. The bounding box marks the detected head, dorsal fin, and tail regions along with their confidence levels.

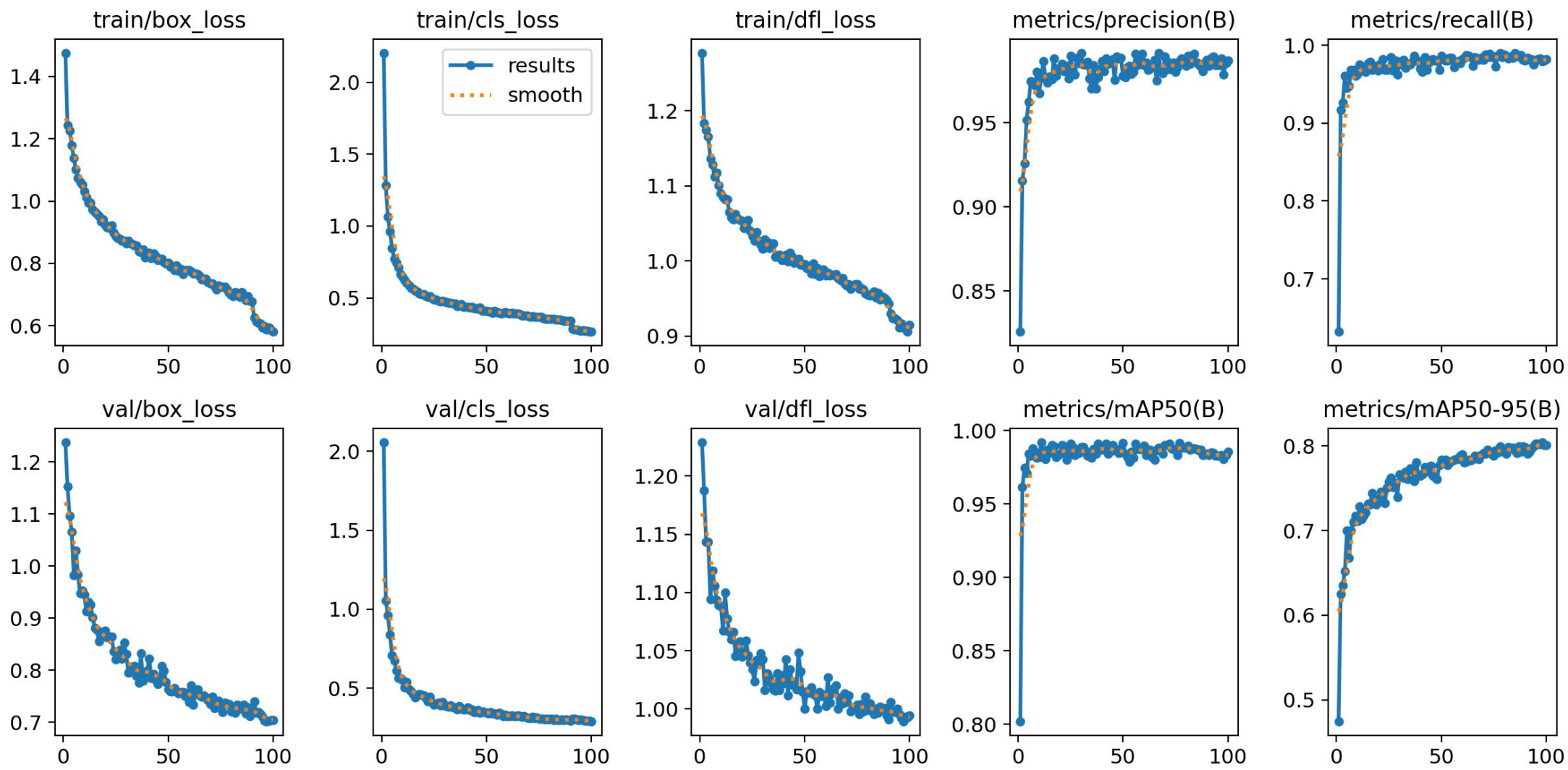

**Fig. 8.** YOLOv8n Training Loss and mAP Curve. The horizontal axis is epoch, and the vertical axis contains box_loss, cls_loss, and mAP50.

### *3.2.2 Results of Slow Path Fusion Classifier*

The slow path fusion classifier is a three layer fully connected network. The input consists of a 8192dimensional fused feature vector obtained by concatenating four parts: the global feature of 2048 dimensions, the head feature of 2048 dimensions, the dorsal fin feature of 2048 dimensions, and the tail feature of 2048 dimensions. The output is 6 categories. The fusion classifier was trained for 100 epochs (Fig. 9). The internal validation accuracy on the 6 confusable species reached 100.00%, indicating that the fusion of global and local features effectively captures subtle morphological differences among closely related species. In the joint system evaluation, on the 4,818image training set, the slow path intervention increased the overall accuracy from 96.72% to 98.94% in Stage 1, and from 99.29% to 100.00% in Stage 2. More importantly, on the independent test set 607 images, the standalone fast path achieved an accuracy of 99.67%, where the 2 misclassified samples were concentrated in closely related species with extremely similar morphologies; after triggering the slow path, all errors were successfully corrected, elevating the overall test accuracy to 100.00%. The overall accuracy comparisons are summarized in Table 7 and Fig. 10.

**Table 7**

Overall system accuracy with and without the slow path.

| Stage | Fast path accuracy | Accuracy after slow path | Improvement |
|---|---|---|---|
| Stage 1 | 96.72% | 98.94% | +2.22% |
| Stage 2 | 99.29% | 100.00% | +0.71% |

Note: The slow path is activated only when the fast path predicts one of the 6 confusable species.

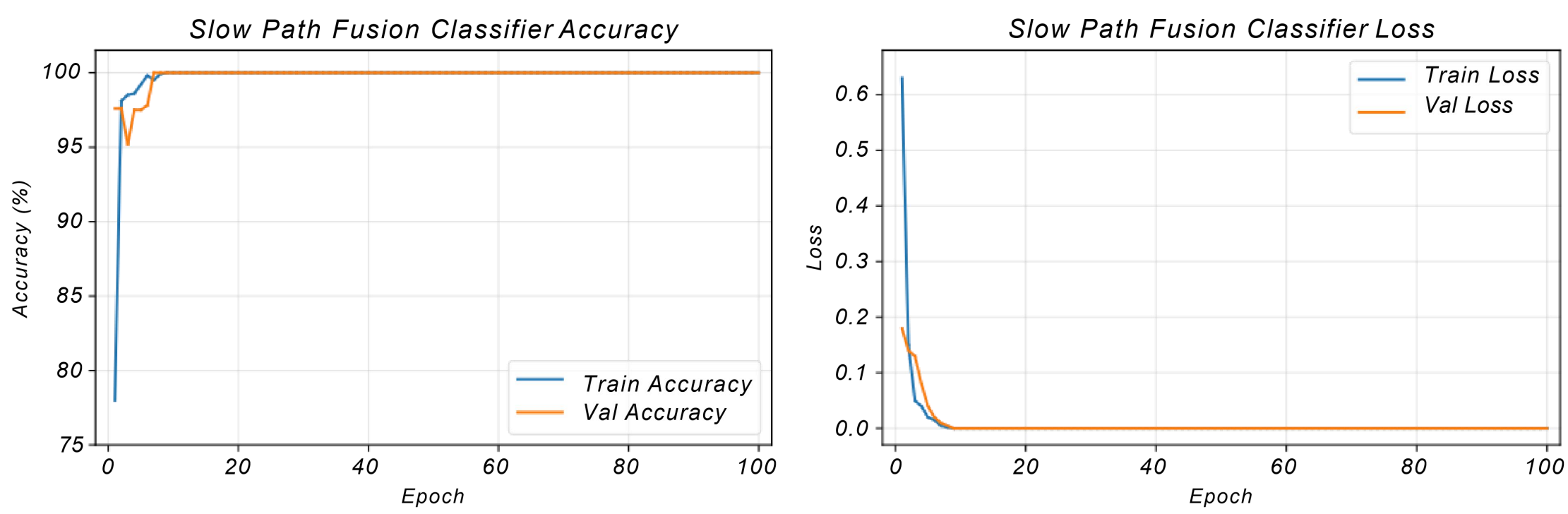

**Fig. 9.** Training curve of the slow path fusion classifier. Training and validation accuracy varies with 100 epochs.

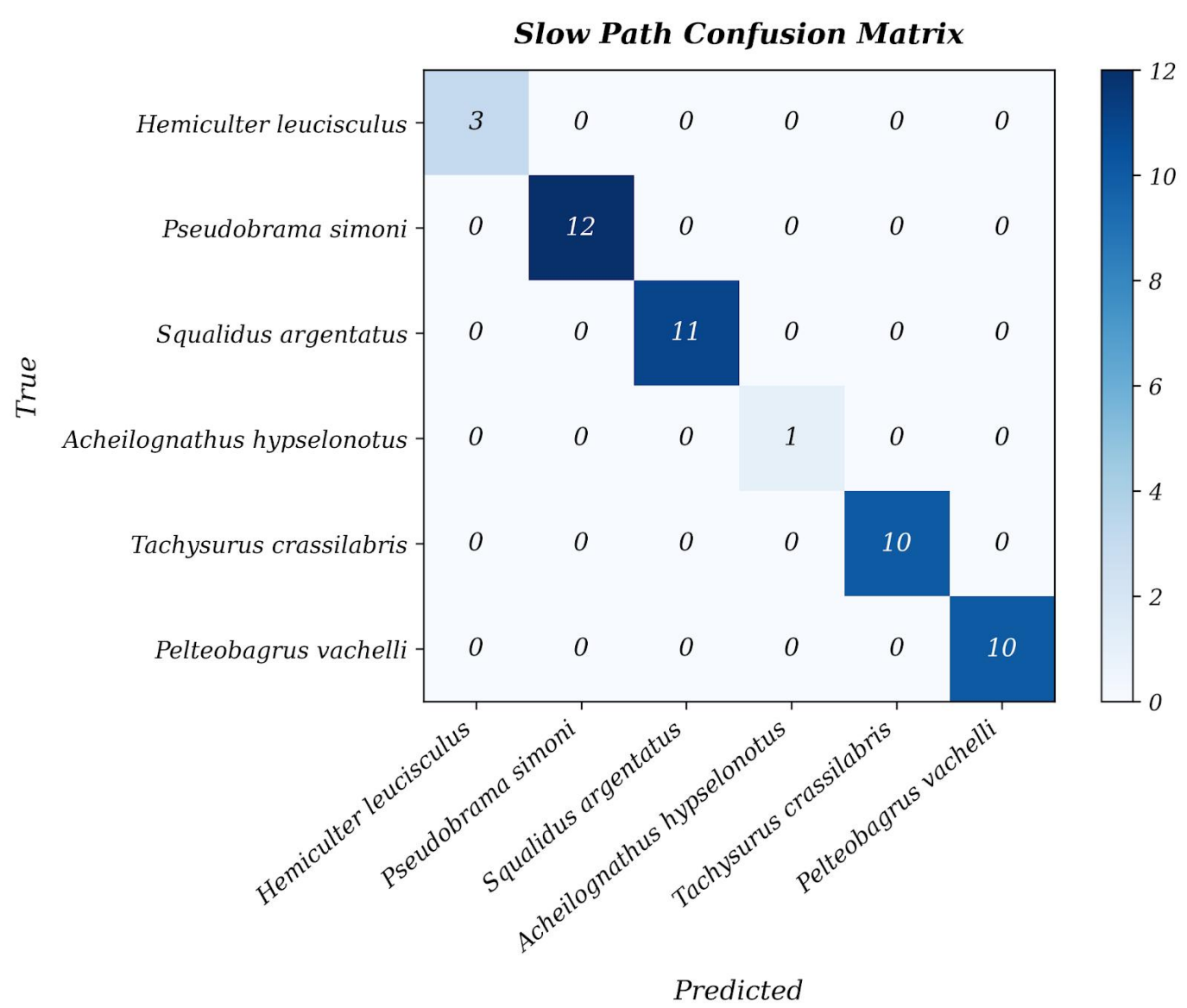

**Fig. 10.** Confusion matrix of the slow path fusion classifier on 6 easily confused fish species.Note:Evaluated on the 47-image internal validation subset of the 6 confusable species. *Pelteobagrus vachelli* in the figure panels refers to the valid taxonomic name *Tachysurus vachellii*.

To further evaluate the computational overhead of the system, we measured the parameters, FLOPs,

and single-module inference time in a CPU environment without GPU acceleration. The results are summarized in Table 8. The fast path comprises 23.61 M parameters with 4.13 G FLOPs and a single-image inference time of 876.34 ms, whereas the slow path consists of global and local ResNet50 encoders alongside a fusion classifier, totaling 51.34 M parameters and an inference time of 2,991.95 ms. For each slow path execution, the global encoder processes one global image while the local encoder handles three regional crops (head, dorsal fin, and tail), resulting in 16.52 G FLOPs. Based on the cascaded latency model formulated in Eq. (4), the empirical trigger proportion $p$ remains small in practice. Specifically, among the 607 test sets, the accuracy of the fast path was 99.67%, with only 2 errors. After the slow path was introduced, all the errors were corrected, resulting in an overall accuracy of 100% (Table 10). Although the standalone fast path achieves high preliminary accuracy on the test set, the introduction of YOLOv8n guided slow path is essential for mission critical ecological monitoring. Unconstrained aquatic environments characterized by severe turbidity, illumination shifts, or partial occlusions, global only fast models are prone to catastrophic failure. The slow path functions as a robust safety net, minimizing classification risks for protected confusable fish species during field monitoring.

**Table 8**

Computational cost analysis of the fast and slow paths.

| Module | Params (M) | FLOPs (G) | Time (ms) | FPS |
|---|---|---|---|---|
| Fast Path | 23.61 | 4.13 | 876.34 | 1.1 |
| Slow Path (total) | 51.34 | 16.52 | 2991.95 | 0.3 |
| Global Encoder | 23.51 | 4.13 | - | - |
| Local Encoder | 23.51 | 12.39 | - | - |
| Fusion Classifier | 4.33 | - | - | - |

Note: The global encoder processes one image, while the local encoder processes three cropped images (head, dorsal fin, and tail). The calculated FLOPs here only account for the encoders of the slow path, and do NOT include the computational overhead from YOLO, SAM3, and the LLM generation module. Total encoder FLOPs is 16.52 G.

### *3.3 LLM Enhancement Results*

The backend uses Qwen3-14B as the base model. To evaluate the effectiveness of the knowledge enhanced module in fish knowledge recognition and generation, parameter efficient finetuning experiments were conducted using Qwen3-14B across 51 fish species (255 evaluation samples). Four configurations were compared under identical knowledge base and data splits: the RAG Baseline (Group A), RAG with soft token Group B, 8 learnable prompt vectors), RAG with LoRA (Group C, rank = 8), and the joint RAG with soft token and LoRA model (Group D, Ours). As shown in Table 9, introducing soft token alone Group B led to a decline in entity recognition, with scientific name accuracy and genus name accuracy dropping to 60.39%, indicating that a small number of input side vectors alone cannot sufficiently alter the model's internal representation without parameter adaptation. In contrast, LoRA fine-tuning Group C effectively enhanced domain adaptability, achieving 93.33% scientific name accuracy and 92.94% Genus name accuracy. Most notably, the joint optimization of soft token and LoRA Group D achieved optimal comprehensive performance: Scientific name accuracy increased from 91.37% Group A to 100.00% (+8.63 percentage points), Genus Accuracy increased from 90.20% to 98.82% (+8.62 percentage points), and ROUGE-L improved from 0.335 to 0.655, while maintaining high completeness (90.07%). Soft token and LoRA exhibit strong complementarity to soft token constrains domain entity relationships via learnable input representations, while LoRA adapts internal attention weights to domain specific phrasing, jointly suppressing biological hallucinations.Beyond enhancing factual accuracy, integrating the fine-tuned LLM with the visual cascade provides structured taxonomic context to support post-recognition verification. Traditional classifiers output only an isolated species name, leaving frontline practitioners without diagnostic reference criteria. In our framework, once the slow pathway resolves confusable species such as *Hemiculter leucisculus* and *Squalidus argentatus* (or *Pseudobrama simoni*), the backend retrieves the corresponding taxonomic records and generates a structured species profile (Fig. 11). Specifically, conditioned on the predicted taxon and retrieved morphological records, the LLM summarizes key diagnostic traits (such as the abdominal keel and lateral line curvature in *H. leucisculus* versus the sub-inferior mouth, short maxillary barbels, and silvery lateral stripe in *S. argentatus*) enabling practitioners to cross-check the visual specimen against authoritative ichthyological descriptions.

**Table 9**

ROUGE-L and Lexical F1 comparison of the backend LLM.

| Metric | Group A | Group D | Absolute improvement | Relative improvement |
|---|---|---|---|---|
| ROUGE-L | 0.335 | 0.655 | 0.320 | 95.5% |
| Lexical-F1 | 0.598 | 0.948 | 0.350 | 58.5% |

Note: Group A uses retrieval augmented generation only. Group D additionally uses soft token and LoRA for parameter efficient fine-tuning.

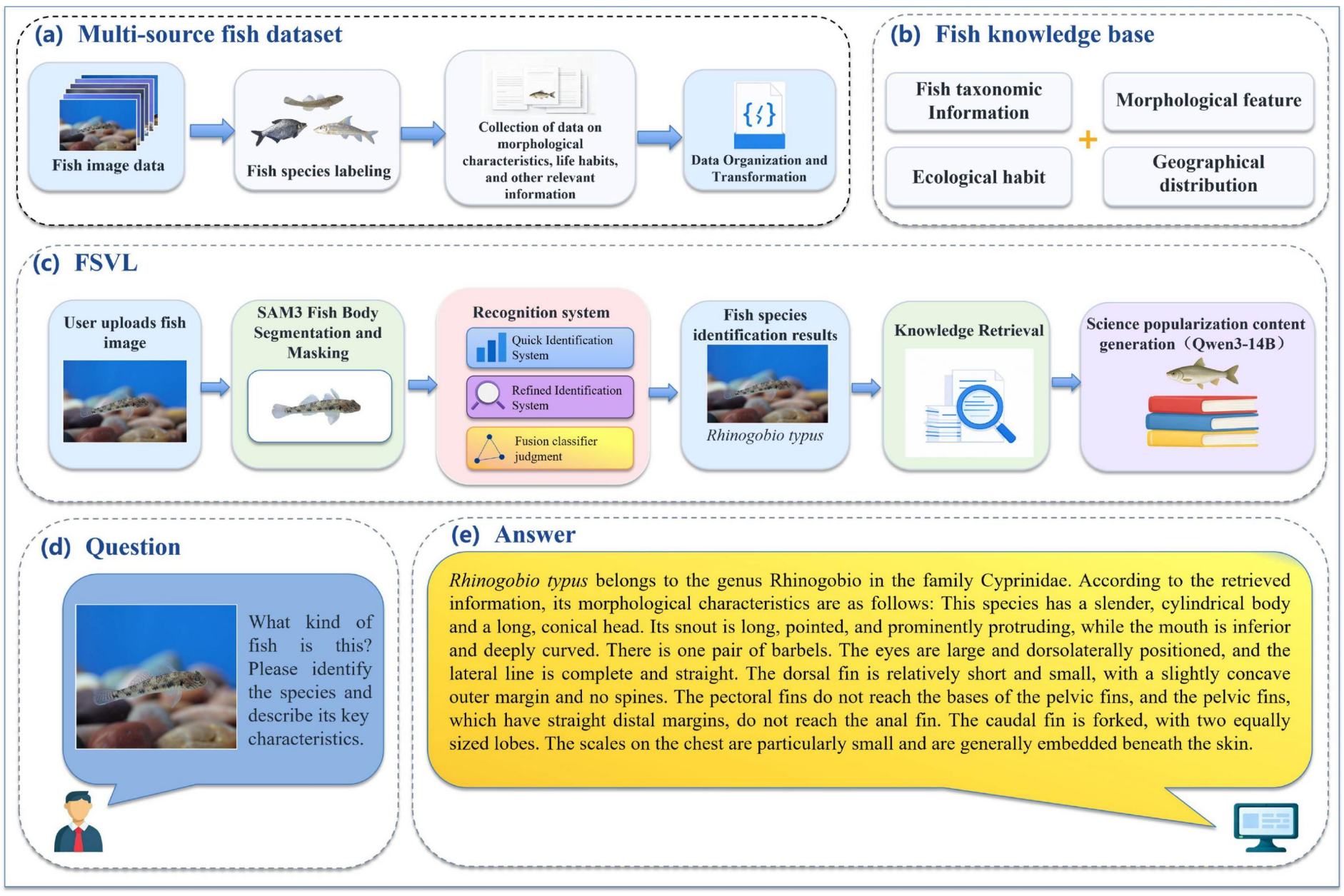

**Fig. 11.** Input and output of the backend LLM.

### *3.4 Ablation Experiment Results*

To verify the effectiveness of the fixed routing strategy proposed in this paper, we compared various routing strategies on a test set consisting of 607 images that cover all 51 types of fish species. All strategies were combined based on the same set of fast and slow path prediction results to ensure a fair comparison. The comparison strategies included only fast paths, only slow paths, triggering slow paths for all cases, triggering slow paths based on confidence threshold values (0.5, 0.7, 0.9), and the fixed routing strategy proposed in this paper. The experimental results are shown in Table 10.

**Table 10**

Ablation results of different routing strategies.

| Routing Strategy | Accuracy (%) | Correct/Total |
|---|---|---|
| Fast Path Only | 99.67 | 605/607 |
| Slow Path Only | 30.15 | 183/607 |
| All Slow | 30.15 | 183/607 |
| Confidence < 0.5 | 99.67 | 605/607 |
| Confidence < 0.7 | 99.84 | 606/607 |
| Confidence < 0.9 | 99.34 | 603/607 |
| Fixed Routing (Ours) | 100 | 607/607 |

As shown in the ablation comparison, applying the specialized expert slow path globally across all 51 classes without fast-path screening yields an accuracy of only 30.15%. This occurs because the slow-path classifier is architecturally tailored to resolve confusable fish taxa rather than act as a standalone 51-class classifier. This result validates the structural necessity of the fast-slow collaborative routing mechanism, confirming that the lightweight fast path is essential for broad preliminary screening. The routing strategies based on confidence threshold have accuracy rates of 99.67%, 99.84%, and 99.34%, respectively, which are all lower than the fixed routing strategy. This shows that relying solely on confidence cannot accurately identify difficult samples. The fixed routing strategy proposed in this paper achieves a robust empirical accuracy of 100.00% on the test set, outperforming comparative threshold-based baselines in this evaluation. This strategy utilizes the prior knowledge of fish classification and limits the triggering conditions of the slow path to 6 types of easily confused closely related species, thereby accurately calling the slow path for correction when necessary, and directly using the fast path results for other easily classified samples, avoiding the slow path's misjudgment of non-target classes.Demonstrating that the fixed routing strategy effectively balances the efficiency of the fast path and the fine-grained recognition ability of the slow path. To further verify the effectiveness of the knowledge enhanced module proposed in this paper in the tasks of fish knowledge recognition and generation, this paper conducted parameter efficient fine tuning ablation experiments on 51 types of fish species and 255 evaluation samples. The experiments set the RAG baseline model soft token knowledge injection model, LoRA parameter efficient finetuning model, and soft token and LoRA joint optimization model. All experiments used the same knowledge base, training data division, and test set to ensure the comparability of the experimental results.

**Table 11**

Ablation results of parameter efficient fine tuning strategies.

| Method | Scientific Accuracy (%) | Genus Accuracy (%) | ROUGE-L | Completeness (%) |
|---|---|---|---|---|
| RAG Baseline | 91.37 | 90.2 | 0.335 | 90.46 |
| RAG with soft token | 60.39 | 60.39 | 0.241 | 89.48 |
| RAG with LoRA | 93.33 | 92.94 | 0.2854 | 90.13 |
| RAG with soft token and LoRA | 100 | 98.82 | 0.655 | 90.07 |

As shown in Table 11, after introducing soft token alone, the model experienced a certain decline in the classification entity recognition indicators. Its Scientific Name Accuracy and Genus Accuracy were 60.39% and 60.39% respectively. This indicates that relying solely on a small number of trainable knowledge vectors is insufficient to significantly change the original knowledge representation of the large language model. A combination of the parameter adaptation mechanism is needed to further optimize the knowledge injection effect. In contrast, the LoRA fine tuning strategy can effectively enhance the model's adaptability to knowledge in the fish domain, achieving scientific name accuracy and genus name accuracy of 93.33% and 92.94% respectively. At the same time, it maintains a high level of text generation quality. After further combining soft token and LoRA, the model achieved the best comprehensive performance. Scientific name accuracy and genus name accuracy reached 100.00% and 98.82%, representing improvements of 8.63 and 8.62 percentage points over the RAG baseline, respectively, while ROUGE-L increased from 0.335 to 0.655 (+0.320). This shows that the combined optimization strategy can effectively enhance the model's ability to understand fish classification knowledge and professional descriptive information. Moreover, soft token and LoRA exhibit certain complementarity. Soft token mainly enhances the model's constraint ability on domain entity relationships by introducing learnable knowledge representations, while LoRA improves the model's adaptability to domain text expression methods by adjusting some parameters. The combination of the two enables the model to simultaneously obtain accurate fish classification information and a more professional description ability that conforms to the content of the knowledge base.

## 4. Discussion

### *4.1 Evaluation of Asymmetric Routing Efficiency and Computational Overhead*

#### *4.1.1 Fast Path Accuracy*

The fast path of this study is based on the ResNet50 backbone network, achieving efficient preliminary classification across 51 fish species in the middle reaches of the Yangtze River. The robust performance achieved is primarily attributed to effective background standardization and data preprocessing. By utilizing advanced image segmentation to remove environmental disturbances such as water bodies and substrates, the model focuses purely on intrinsic fish morphology rather than extraneous environmental textures, thereby mitigating domain shifts induced by complex aquatic environments. This highlights the critical role of data quality and standardization in field-deployable vision systems. Compared with existing aquatic recognition research, the fast path maintains high recognition capability even across a large number of categories and closely related species, verifying the effectiveness of the employed data preprocessing strategies and confirming that high-quality visual inputs are fundamental to overcoming the performance bottlenecks of fine-grained aquatic categorization (Kofidis et al., 2026).

#### *4.1.2 Routing Arbiter and Computing Power Allocation*

The core innovation of the fast-slow dual path architecture proposed in this study lies in the dynamic routing that enables on demand allocation of computing power. Traditional single models often require performing equally complex inferences on all samples, resulting in wasted computational resources or compromised accuracy when confronting fine-grained variations. This system assigns 51 fish species to the fast path for rapid preliminary screening, and only triggers the slow path for refined secondary identification when prediction results fall within predefined confusable taxa. This design draws on the idea of hierarchical decision making and aligns with the strategy of coarse screening followed by precise identification commonly adopted in biological taxonomy. The routing arbiter determines whether to activate the slow path based on the predicted category of cost-effective fast screening without relying on arbitrary confidence thresholds. The slow path functions as a robust fallback mechanism to correct residual errors, helping to reduce misclassification among confusable fish species on the evaluated dataset. From the perspective of computing power allocation, while the dynamic routing mechanism restricts the computationally intensive slow path primarily to ambiguous or confusable samples, the precise routing trigger proportion in unconstrained field scenarios requires comprehensive empirical profiling. Consequently, the actual system-level average latency depends directly on task-specific ambiguity distributions rather than defaulting to lightweight performance. This framework provides a scalable resource allocation architecture designed to balance screening throughput with fine-grained precision for ecological monitoring tasks (Jang et al., 2025).

### *4.2 Discriminative Efficacy of Taxonomically Grounded Global-Local Fusion*

#### *4.2.1 Local Feature Recognition Method*

In the identification of closely related fish species, relying solely on global features is insufficient to distinguish species with highly similar appearances due to shared body proportions and colorations. Therefore, local feature recognition plays a crucial role in identifying difficult samples, as local attributes possess strong discriminative power in fine-grained classification (You et al., 2026). Prior studies have demonstrated that local feature enhancement improves classification performance under unconstrained conditions such as partial occlusion, varying postures, or developmental stage shifts (Zhai et al., 2023). In field monitoring scenarios, fish specimens are frequently captured under imperfect angles or suboptimal lighting, making targeted local feature extraction vital for reliable identification. This study selects the head, dorsal fin, and tail as local feature extraction regions based on core evolutionary and taxonomic criteria. The object detector accurately localizes these key anatomical areas, providing structured and salient inputs for subsequent feature fusion, thereby bridging computer vision representations with human expert inspection logic (Zhai et al., 2023; You et al., 2026).

#### *4.2.2 Fusion of Local Features and Global Features*

Based on the identification of local features, further integration of global features can enhance the recognition accuracy. Global features provide the overall outline, body proportion, and color information of the fish, while local features focus on the subtle morphological differences in key areas. The two complement each other, effectively capturing the subtle differences between closely related species. In this study, global features are combined with three local features to form a fused vector, which is sent to a fully connected classifier. Experimental results show that the fused classifier achieved a high internal validation accuracy for closely related species and further improved the overall accuracy of the system after the slow path intervention on the complete test set, with very few error samples. This result is consistent with existing research (You et al., 2026). GCCR achieves optimal performance through adaptive weight fusion to balance global and local features adaptively, and has achieved the best performance on multiple finegrained datasets. GLHPM explicitly combines global and local features, clearly indicating that the global structure provides semantic guidance, and the local details refine the global representation, performing exceptionally well in few sample fine-grained tasks (Zhang et al., 2026). Compared with similar studies using attention mechanisms or dual branch networks, this study is driven by fish taxonomy knowledge, has a simpler structure, and is more easily implemented in engineering. Experimental results show that the strategy of fusing global and local features has high reliability in fine-grained identification of closely related fish species and can effectively improve the overall accuracy of the system (You et al., 2026, Zhang et al., 2026).

### *4.3 Efficacy of Parameter-Efficient Adaptation in Grounded Ecological Reasoning*

In the backend knowledge generation process, this study compared schemes using retrieval augmented generation alone with those incorporating parameter efficient fine-tuning strategies. Experimental results indicate that joint adaptation significantly outperforms the retrieval baseline in scientific and genus accuracy, confirming that parameter efficient fine-tuning substantially improves domain specific generation quality. While retrieval mechanisms provide external factual grounding, language models can still experience semantic deviations, terminology omissions, or structural inconsistencies without targeted adaptation (Jang et al., 2025). FMRAG is the first MLLM with RAG framework for fisheries, which significantly improves the recognition performance of rare species by retrieving visual similar fish images and text records, proving the effectiveness of RAG for the long tail distribution (Asai et al., 2024, Song et al., 2026). Taxonomic Reasoning for Rare Arthropods combines image descriptions with RAG for rare arthropod classification and reveals the rule that the quality of the knowledge base is more important than the retrieval strategy. BioChat adopts RAG combined with routing mechanisms in the biodiversity question answering system, effectively improving the factual accuracy and reducing the illusion rate (Jang et al., 2025). Studies show that although RAG retrieval can provide accurate knowledge sources for large models, the model may still encounter problems such as term omissions and semantic deviations when organizing language. After introducing soft token and LoRA for efficient parameter fine tuning, the content generated by the model significantly improves in terms of semantic consistency and lexical accuracy. Soft Token learns continuous prompt representations related to the fish knowledge generation task, while LoRA adapts the internal attention weights of the model in a lightweight manner. Together, they achieve joint parameter efficient adaptation of the input representation space and the model parameter space. RAG is responsible for providing factual basis, while soft token and LoRA are responsible for constraining the generation behavior, effectively suppressing the illusion problem. More crucially, this joint RAG and LLM enhancement addresses a fundamental limitation in ecological AI: the lack of interpretability. By synthesizing the predicted taxonomic labels with RAG, the system produces structured biological profiles that bridge automated classification with domain-specific reference knowledge. Rather than directly inspecting image pixels or independently verifying specimen-specific diagnostic traits from bounding boxes, the language backend conditions on the predicted taxon and retrieved records to summarize diagnostic characteristics, ecological habits, and conservation status. This provides frontline fishery administrators and conservationists with accessible taxonomic reference information for secondary verification. Current biological recognition systems generally only perform classification and do not output knowledge. This study connects the classification results with the RAG knowledge base and constrains the generation through soft token and LoRA, achieving a closed loop from machine classification to knowledge generation. Compared with the scheme using only RAG, this study proves the necessity of efficient parameter fine tuning in domain generation tasks (Jang et al., 2025, Song et al., 2026). Beyond algorithmic innovations, this comprehensive framework directly addresses the operational demands of the Yangtze River's Ten-Year Fishing Ban. As the moratorium facilitates the gradual recovery of rare, endangered, and closely related species, real time and harm free biodiversity tracking becomes critical. The FSVL system, tailored for edge deployment, enables in situ nonintrusive monitoring without disrupting aquatic habitats. Furthermore, the structured knowledge generated by the RAG LLM backend empowers fishery administration personnel who may lack formal taxonomic training with immediate, expert level biological insights. By explicitly outlining diagnostic traits, ecological habitats, and conservation status, the system significantly lowers the barrier to precise law enforcement, prevents misidentification of protected species, and provides an interactive platform for public science education. Thus, our approach transcends traditional image classification, evolving into an integrative ecological management asset.

*4.4 Current Methodological Bottlenecks and Future Research Trajectories*

Current research on fine-grained visual recognition and RAG still faces several common challenges. Training data for closely related species are often scarce and imbalanced, making models susceptible to overfitting on majority classes with inadequate generalization to low frequency minority classes under long tail distributions. Domain shifts induced by varying water bodies, lighting conditions, and distinct acquisition devices further degrade model stability in complex field scenarios. Concurrently, the multi branch structure for fusing global contours and local details introduces non-negligible computational overhead, while cascaded reasoning increases end to end latency. These constraints impede real time deployment on compute and power limited edge monitoring terminals. Additionally, while RAG incorporates external knowledge, static repositories and rigid routing thresholds risk triggering false negatives. The field also lacks standardized evaluation protocols to systematically assess professional terminology consistency and biological hallucination rates. Specifically, in the current FSVL implementation, the slow-path specialist covers six representative confusable species, and the language backend generates species profiles conditioned on the predicted taxon and retrieved records rather than directly verifying pixel-level morphological traits. Moreover, end-to-end latency profiling and external validation across unconstrained underwater environments remain to be conducted in future iterations.

To address these challenges, future research can advance along several trajectories. Regarding data and generalization, domain adaptation and few shot learning strategies can be integrated to reduce reliance on large scale annotated data, enhancing cross domain robustness across diverse aquatic environments on broader public datasets. In terms of computational efficiency and deployment, model pruning, quantization, and knowledge distillation warrant investigation to compress inference overhead while maintaining accuracy, coupling lightweight backbones with edge computing architectures for real time field operation. Regarding knowledge augmentation and evaluation, constructing dynamic vector databases with multimodal hybrid retrieval and continuous learning will improve knowledge coverage and timeliness. Meanwhile, establishing standardized benchmarks and evaluation protocols supplemented with metrics such as field accuracy, intragenus discriminability, and hallucination penalties will yield a more rigorous evaluative framework.

## 5. Conclusion

This study proposes and implements a knowledge enhanced fast slow dual path collaborative architecture to address the challenges of fine-grained identification and reliable knowledge generation for closely related fish species in the Yangtze River basin. To balance inference efficiency and classification accuracy, the system employs a dynamic routing mechanism. A lightweight fast path performs efficient coarse screening, while a slow path is adaptively triggered for confusable species. By integrating expert taxonomic knowledge, the slow path locates key local morphological regions and utilizes a standardized global local feature fusion strategy to significantly improve fine-grained distinguish ability. Furthermore, to overcome the factual hallucinations commonly found in LLMs, the backend integrates a RAG framework. By applying a joint soft token and LoRA adaptation strategy with a masked response only loss, the system effectively constrains the model to generate accurate and structured scientific content based on retrieved facts. In conclusion, this research successfully bridges the gap between expert taxonomic experience and deep learning architectures, completing a seamless pipeline from visual perception to intelligent knowledge generation. By establishing a robust, non-destructive paradigm for biodiversity monitoring and automated science popularization, the proposed system demonstrates profound practical value in supporting the ongoing Ten-Year Fishing Ban in the Yangtze River, paving the way for next generation intelligent ecological management and sustainable fisheries conservation.

## Acknowledgments

The authors would like to thank all individuals who contributed to this study. We are especially grateful to Prof. Qiwei Wei from the Yangtze River Fisheries Research Institute and Dr. Lei Li from the Heilongjiang River Fisheries Research Institute, both affiliated with the Chinese Academy of Fishery Sciences, for their valuable guidance on high-quality live fish photography methodologies. We also thank

Dr. Xiao Xie from Hunan University of Arts and Science for assistance in fish specimen collection and photography. In addition, we sincerely appreciate the undergraduate students from the College of Fisheries at Southwest University, including Ming Fu, Yile Sun, Xiaoxuan Wang, Weixi Shi, and Liyun Wang, for their diligent assistance in data annotation.

This work was supported by National Key R&D Program of China（Grant No. 2025YFD2400100） and Central Public-interest Scientific Institution Basal Research Fund, CAFS (NO.2026TD104).

**CRediT authorship contribution statement**

**Meng Liang:** Project administration, Resources, Data curation, Conceptualization, Methodology, Formal analysis, Software, Visualization, Investigation, Writing - original draft, Writing - review & editing. **Fucheng Wang:** Supervision, Resources, Conceptualization, Software, Formal analysis, Visualization, Writing - review & editing. **Lang Zhao:** Methodology, Formal analysis, Visualization, Writing - review & editing. **Jun Chen:** Software, Investigation, Writing - review & editing, Data curation. **Ziying Li:** Investigation, Writing - review & editing. **Yuxiang Zhao:** Visualization, Writing - review & editing. **Haozhuang Chi:** Methodology, Investigation, Writing - review & editing. **Wenkai Fang:** Investigation, Writing - review & editing. **Qiong Zhou:** Data curation. **Junlin Yang:** Writing - review & editing. **Jinming Wu:** Conceptualization, Project administration, Funding acquisition, Supervision, Writing - review & editing.

**Declaration of competing interest**

The authors declare that they have no known competing financial interests or personal relationships that could have appeared to influence the work reported in this paper.

**Data and code availability**

The model implementation code and evaluation scripts used in this study are available from the corresponding author upon reasonable academic request. The raw fish specimen images are currently subject to publication restrictions related to an upcoming illustrated taxonomic atlas and may be shared on a limited basis for non-commercial academic validation upon reasonable request and under appropriate data-use agreements.